\documentclass[dvipsnames]{article}

\usepackage{microtype}
\usepackage{graphicx}
\usepackage{subfigure}
\usepackage{booktabs} 
\usepackage{threeparttable} 
\usepackage{multirow} 
\usepackage{fancyvrb} 
\usepackage{fvextra}
\usepackage{tcolorbox}
\tcbuselibrary{listings,breakable}  

\newtcblisting{jsonschema}{
  listing only,
  colback=gray!5!white,
  colframe=gray!80!black,
  title=JSON Schema,
  breakable,
  fontupper=\ttfamily\footnotesize,
  left=0pt,right=0pt,top=2pt,bottom=2pt,
}
\definecolor{Mycolor}{HTML}{00F9DE}

\usepackage{hyperref}

\usepackage[accepted]{icml2025}

\usepackage{amsmath}
\usepackage{amssymb}
\usepackage{mathtools}
\usepackage{amsthm}

\usepackage[capitalize,noabbrev]{cleveref}

\theoremstyle{plain}

\theoremstyle{definition}

\theoremstyle{remark}

\usepackage[textsize=tiny]{todonotes}

\icmltitlerunning{Are LLM Belief Updates Consistent with Bayes' Theorem?}

\begin{document}

\twocolumn[
\icmltitle{Are LLM Belief Updates Consistent with Bayes' Theorem?}



\icmlsetsymbol{equal}{*}

\begin{icmlauthorlist}
\icmlauthor{Sohaib Imran}{lec,scc}
\icmlauthor{Ihor Kendiukhov}{utcs}
\icmlauthor{Matthew Broerman}{}
\icmlauthor{Aditya Thomas}{ir}
\icmlauthor{Riccardo Campanella}{nsu}
\icmlauthor{Rob Lamb}{lec,jba}
\icmlauthor{Peter M. Atkinson}{lec,ges,cst}
\end{icmlauthorlist}

\icmlaffiliation{lec}{Lancaster Environment Centre, Lancaster University, Lancaster LA1 4YQ, UK}
\icmlaffiliation{scc}{School of Computing and Communications, Lancaster University, Lancaster LA1 4WA, UK}
\icmlaffiliation{jba}{JBA Trust, 1 Broughton Park, Skipton BD23 3FD, UK}
\icmlaffiliation{ges}{Geography and Environmental Science, University of Southampton, Highfield, Southampton SO17 1BJ, UK}
\icmlaffiliation{cst}{College of Surveying and Geo-Informatics, Tongji University, No.1239, Siping Road, Shanghai, PR China, 200092}
\icmlaffiliation{utcs}{University of Tuebingen, Department of Computer Science, Sand 14, 72076 Tuebingen, Germany}
\icmlaffiliation{nsu}{Graduate School of Natural Sciences, Faculty of Science, Utrecht University, Heidelberglaan 8, 3584 CS Utrecht, The Netherlands}
\icmlaffiliation{ir}{Independent Researcher}

\icmlcorrespondingauthor{Sohaib Imran}{s.imran1@lancaster.ac.uk}


\icmlkeywords{Coherence, Consistency checks, Language Models, LLMs, Bayes' theorem}

\vskip 0.3in
]



\printAffiliationsAndNotice{\icmlEqualContribution} 

\begin{abstract}
%
Do larger and more capable language models learn to update their "beliefs" about propositions more consistently with Bayes' theorem when presented with evidence in-context? To test this, we formulate a Bayesian Coherence Coefficient (BCC) metric and generate a dataset with which to measure the BCC. We measure BCC for multiple pre-trained-only language models across five model families, comparing against the number of model parameters, the amount of training data, and model scores on common benchmarks. Our results provide evidence for our hypothesis that larger and more capable pre-trained language models assign credences that are more coherent with Bayes' theorem. These results have important implications for our understanding and governance of LLMs.
\end{abstract}

\section{Introduction} \label{sec: intro}

Bayes' theorem allows optimally updating credences as observations provide evidence to existing beliefs \citep{lin_bayesian_2024}. Knowing whether large language models (LLMs) internally perform belief updates that approximate this rule is important for understanding and governing their behavior. 

While previous research showed that transformers trained on hidden Markov model data naturally implement a form of constrained Bayesian inference for next token prediction \citep{piotrowski_constrained_2025}, LLMs showed no improvement in Bayesian consistency over abstract propositions rather than tokens \citep{fluri_evaluating_2023}.  However, larger LLMs have been shown to violate other logical and probabilistic axioms less often than smaller models \citep{fluri_evaluating_2023, paleka_consistency_2025}. \citet{mazeika_utility_2025} additionally demonstrated that larger models exhibit higher preference coherence, measured by fewer transitivity violations. They also found that larger models are more decisive and consistent over their preferences, which they interpret as proxies for preference completeness.

If we expect LLMs trained and deployed in the future to be coherent Bayesian updaters, this has profound implications. More coherent agents are more understandable and predictable by agents with similar world models, which translates to greater reliability and robustness, and therefore steerability of LLMs. This also means that human behavior should be more understandable to more coherent LLMs. The latter should make it easier to convey information and specify complex goals (e.g., to empower humans).  On the other hand, it makes it more difficult to conceal information when interacting with LLMs. This makes it difficult to evaluate LLMs without them conditioning on the fact that they are being evaluated \citep{fan_evaluation_2025, needham_large_2025}. 

In partially observable environments, Bayesian belief updating allows for optimal control \citep{astrom_optimal_1965, sondik_optimal_1978}. Agents that additionally have coherent preferences are well modeled as expected utility maximizers (EUMs) \citep{hammond_consequentialist_1988}, which brings many benefits but also serious risks. In particular, if the preferences of EUMs are misaligned with human preferences, they may optimize for world states that humans dislike \citep{everitt_alignment_2018}. EUMs may also resist shutdown or modification of their goals and preferences (i.e. they may be incorrigible) and therefore engage in deception and power seeking \citep{soares_corrigibility_2015, everitt_alignment_2018, hubinger_risks_2021}.

Motivated by the above considerations, and the emerging debate about whether LLMs can reason \cite{shojaee_illusion_2025, opus_comment_2025}, we investigate whether LLMs update their credences in-context over propositions in a manner consistent with Bayes' rule, and how this property scales with model size and capability. The contributions of our research are:

\begin{itemize}
    \item We introduce a novel metric and dataset that compares the observed and expected updates that a model performs across propositions and evidences in multiple conversational contexts.
    \item We demonstrate empirically that larger and more capable LLMs update their credences over propositions in a manner more consistent with Bayes' rule.
    \item We discuss the implications of our results for AI safety and alignment.
\end{itemize}

\section{Bayesian Coherence Coefficient}

For this analysis, the propositions under consideration are a set of classes \textbf{\(  \textbf{C}\)}. Given some evidence \(x\), Bayes' theorem can be used to update one's credences:

\begin{equation}
    P(c | x ) = \frac{P(x | c)P(c)}{\sum_{c' \in \mathbf{C}}P(x|c')P(c')}
\end{equation}
Since the set of all possible classes that some evidence \(x\) points to is infinite, we consider ratios:

\begin{equation}
        \frac {P(c_1 | x )}{P(c_2 | x )} = \frac{P(x | c_1)P(c_1)}{P(x|c_2)P(c_2)}
\end{equation}
where \(c_1, c_2 \in \ \textbf{C}\) are pairs of classes.

We introduce the Bayesian Coherence Coefficient (BCC), which measures the correlation between the expected and observed updates given any evidence, as measured by the log likelihood ratios and the log odds updates, respectively:
\begin{equation} \label{eq: BCC}
\begin{aligned}
\text{BCC}(\theta, \mathcal{D}) = \text{Corr}\Big(&\Delta_{\text{expected}}, \Delta_{\text{observed}}\Big)
\end{aligned}
\end{equation}
Where \(\theta\) is the model being evaluated, and \(\mathcal{D}\) is a dataset composed of multiple categories \(k\) of classes \(c\), evidences \(x\), and conversation histories \(h\). 

The expected and observed updates are:
\begin{equation} \label{eq: expected_update}
    \begin{aligned}
    \Delta_{\text{expected}} &= \text{log likelihood ratio} = \log\frac{P_\theta(x|c_1,h, k)}{P_\theta(x|c_2,h, k)}
    \end{aligned}
    \end{equation}
\begin{equation} \label{eq: observed_update}
\begin{aligned}
\Delta_{\text{observed}} &= \text{log odds update} \\
&=\text{log posterior ratio} - \text{log prior ratio} \\
&= \log\frac{P_\theta(c_1|x,h, k)}{P_\theta(c_2|x,h, k)} - \log\frac{P_\theta(c_1|h, k)}{P_\theta(c_2|h, k)}
\end{aligned}
\end{equation}

where \(P_\theta\) is the LLM-assigned cumulative conditional probability assigned to the tokens composing the class \(c\) or evidence \(x\).

\section{Data and Methodology}

To generate the dataset \(\mathcal{D}\), we prompt an LLM (we use the GPT-4o model accessed via the ChatGPT interface) with a JSON schema, desiderata for the classes, evidences and histories, an example dataset and a category \(k\). The exact prompt including the desiderata is given in Appendix \ref{app-prompt}. We repeat this process for 10 manually curated categories, resulting in a dataset with each category containing at least five classes \(c\), 20 evidences \(x\) and three conversation histories \(h\), as well as class elicitation and evidence elicitation strings, prefixed to encourage the model to generate the classes and evidences, respectively. 

\begin{figure}
  \centering
  \includegraphics[width=\columnwidth]{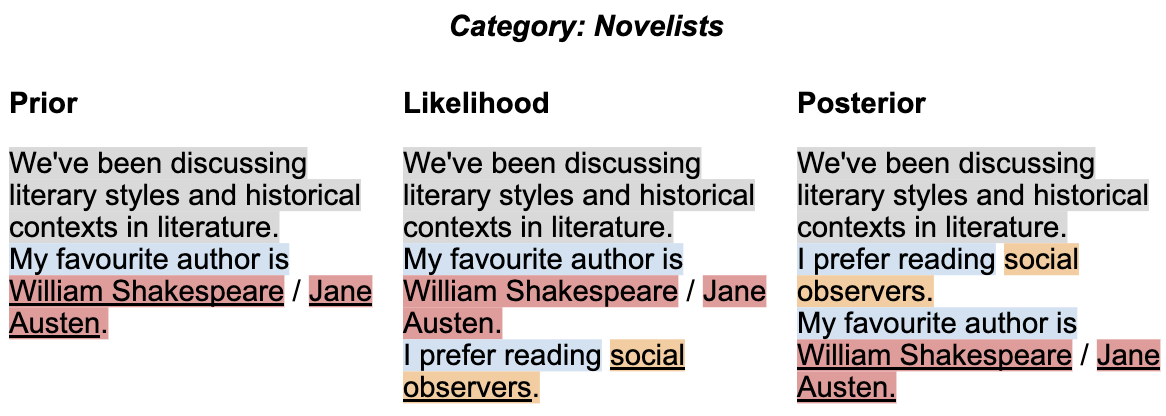}
  \caption{We compute priors, likelihoods and posteriors for all class (red), evidence (orange), history (grey) and category combinations, as the cumulative log probabilities assigned to the underlined tokens, conditional upon the preceding text. Elicitation texts (blue), fixed for each category, are used to encourage the class and evidence tokens. The Bayesian Coherence Coefficient is computed as the correlation between the expected and observed updates for all class pairs (either side of /) within each category and given each evidence and conversation history.}
  \label{fig:data_example}
\end{figure}

Figure \ref{fig:data_example} illustrates how the aforementioned components are used to compute priors \(P_{\theta}(c \mid h, k)\), likelihoods \(P_{\theta}(x\mid c,h, k)\) and posteriors \(P_{\theta}(c \mid x,h, k)\) assigned by the model \(\theta\) being evaluated. Importantly, we use separate instances of the model to compute each of these.

Following the above, for every class pair \((c_1, c_2)\) in each category \(k\), we compute the prior ratio, likelihood ratio and posterior ratio as detailed in equations \ref{eq: expected_update} and \ref{eq: observed_update}, finally computing the BCC as the correlation between all the expected and observed updates (equation \ref{eq: BCC}).

All models were evaluated with the temperature parameter set to 1, which is the temperature that models are typically trained with. Model parameter counts and benchmark scores were obtained from the Open LLM Leaderboard 2 \citep{clementine_fourrier_open_nodate}, which provides standardized evaluations across multiple language model families. Both the parameter counts and benchmark scores reported by the leaderboard may differ slightly from those in official model documentation.

\section{Results}

\begin{figure}
    \centering
    \includegraphics[width=0.98\columnwidth]{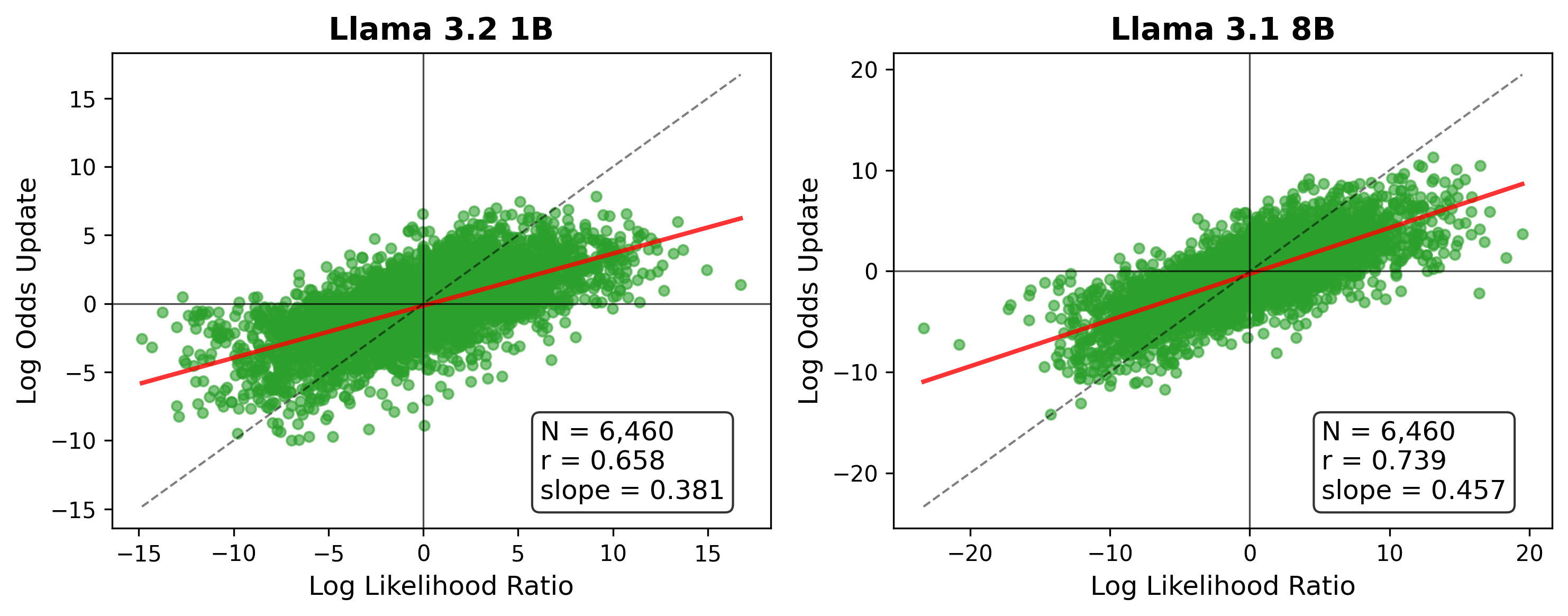}
    \caption{Scatterplots showing the observed updates (log odds updates) against the expected updates (log likelihood ratios) for the Llama 3.2 1B and 3.1 8B models. Each point represents a (class pair, evidence, history, category) tuple from the dataset. The BCC of the model is the correlation (\textit{r} value) between the expected and observed updates. \textit{p} values were too small to be properly rendered and were therefore skipped from this figure. The solid red and dashed black diagonal lines show the observed and ideal update gradients between the observed and expected updates, respectively.}
    \label{fig:llama_comparison}
\end{figure}

\begin{figure}
    \centering
    \includegraphics[width=\columnwidth]{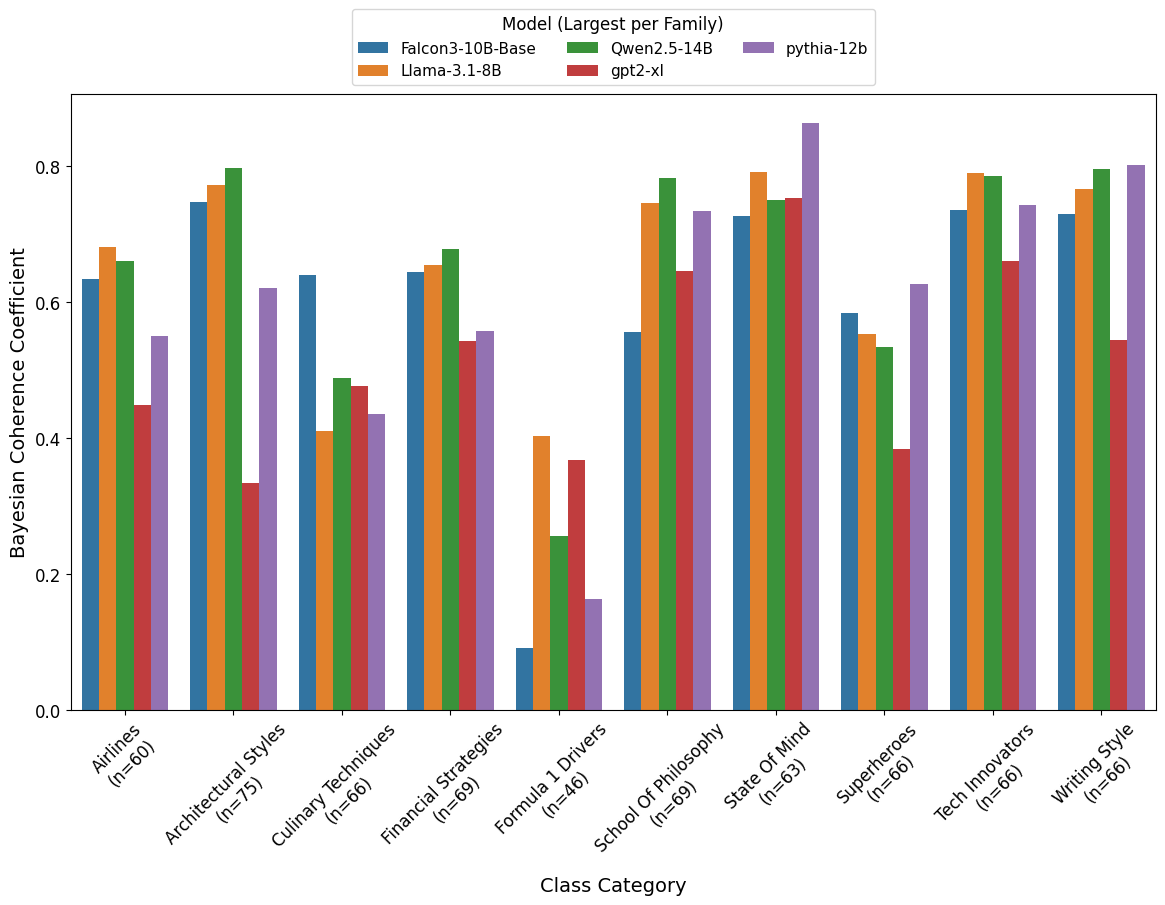}
    \caption{BCC for selected models across the categories in our dataset.}
    \label{fig:bcc_vs_categories}
\end{figure}

We compute the Bayesian Coherence Coefficient (BCC) as described in equation \ref{eq: BCC} across the entire dataset \(\mathcal{D}\), totaling 6460 \((c_1, c_2, x, h, k)\) tuples and for multiple models across five model families. Figure \ref{fig:llama_comparison} illustrates the BCC by visualizing the correlation between expected and observed credence updates for two models from the Llama family. For the largest tested models in each family, we find that BCC is similarly distributed across categories (Figure \ref{fig:bcc_vs_categories}).

\begin{figure}
    \centering
    \includegraphics[width=0.98\linewidth]{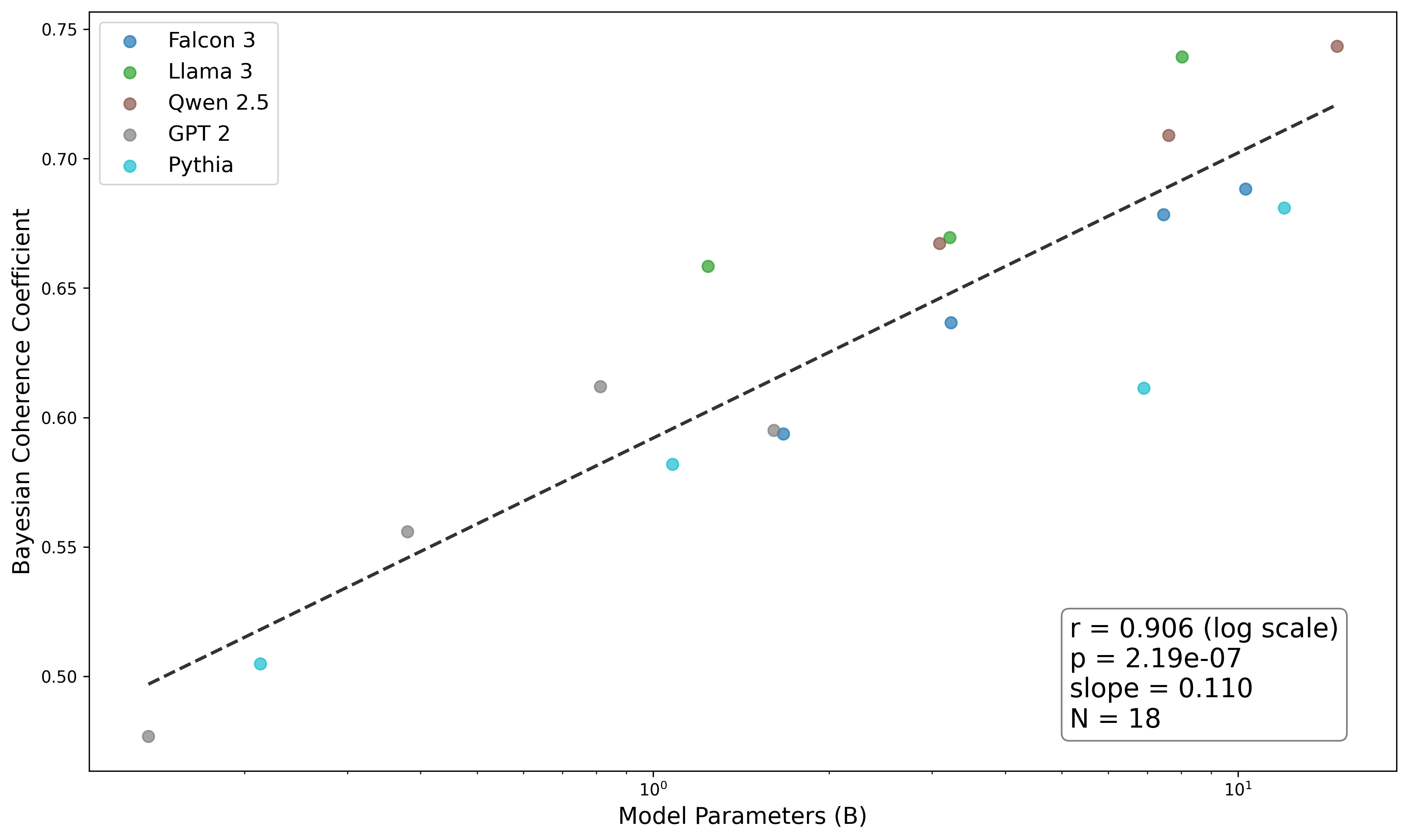}
    \caption{Bayesian Coherence Coefficient as a function of number of model parameters. Each point represents a pre-trained model evaluated on the full dataset. The \textit{x}-axis represents the number of parameters (in Billions) on a logarithmic scale.}
    \label{fig: size}
\end{figure}

All tested models have BCC values greater than 0, implying credence updates more consistent with Bayes' rule compared to a random policy. For all model families tested, BCC increases with scale, with the only exception being the GPT-2 XL and GPT-2 Large models from the GPT-2 family (figure \ref{fig: size} and appendix figure \ref{fig:log_odds_likelihood}). We found a significant positive correlation between BCC and the log of model parameter counts in the models we tested ((\(r = 0.906, p < 10^{-6}\))). Larger models also show a larger proportion of observed and expected updates in the same direction as measured by the Direction Agreement in Table \ref{tab:model_extremes}. 

For all tested models, the gradient of observed vs. expected updates is less than 1. Further investigation reveals that this update gradient is inversely proportional to the negative evidence log likelihood averaged over the class pair (see appendix figure \ref{fig:bcc_vs_avg_evidence_log_likelihood}). Larger models show an update gradient closer to 1 (see table \ref{tab:model_extremes}), with the only exceptions being the GPT-2 Large and XL, and the Pythia 160M and 1B model pairs (see appendix figure \ref{fig:log_odds_likelihood}). Perfect Bayesian updating would entail observed and expected updates being equal; therefore, yielding a gradient of 1.

\begin{figure}
    \centering
    \includegraphics[width=0.98\linewidth]{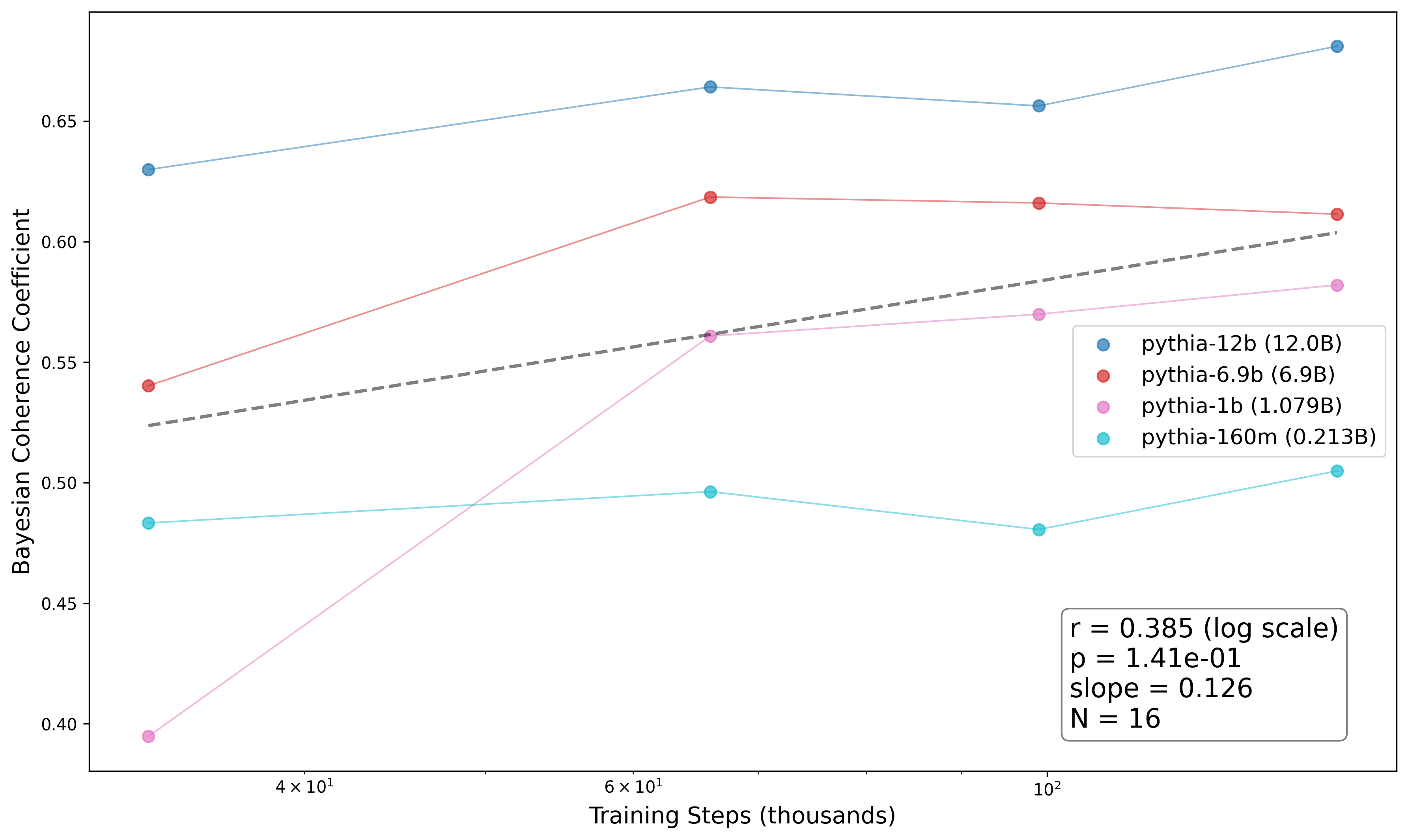}
    \caption{Evolution of the BCC during training for four models with different numbers of parameters (see legend) from the Pythia model family. Each point represents a pre-trained model evaluated on the full dataset. The \textit{x}-axis represents the number of training steps (in thousands) on a logarithmic scale, each of which is over a batch of 2 million tokens.}
    \label{fig:training-data}
\end{figure}

We further use the Pythia model family to evaluate the relationship between BCC and training steps, since these models are available across multiple training checkpoints \citep{biderman_pythia_2023}. We find a non-significant positive trend between BCC and an increasing number of training steps (Figure \ref{fig:training-data}).  

Lastly, we explore how BCC varies with scores obtained by the models on a set of benchmarks commonly used to evaluate model capability, namely, BIG-Bench Hard \citep{suzgun_challenging_2022}, GPQA \citep{rein_gpqa_2023}, MMLU-PRO \citep{wang_mmlu-pro_2024}, IFEval \citep{zhou_instruction-following_2023}, MUSR \citep{sprague_musr_2024}, and Math Lvl 5 (i.e., the level 5 (most difficult) subset of the MATH dataset) \cite{hendrycks_measuring_2021}. We find significant positive correlations (\(p < 10^{-2}\)) between BCC and benchmark score for four of the six tested benchmarks - BIG-Bench Hard, GPQA, MMLU-PRO and Math Lvl 5 - and non-significant positive relationships with the IFEval and MUSR benchmarks.

\begin{figure*}
    \centering
    \includegraphics[width=\textwidth]{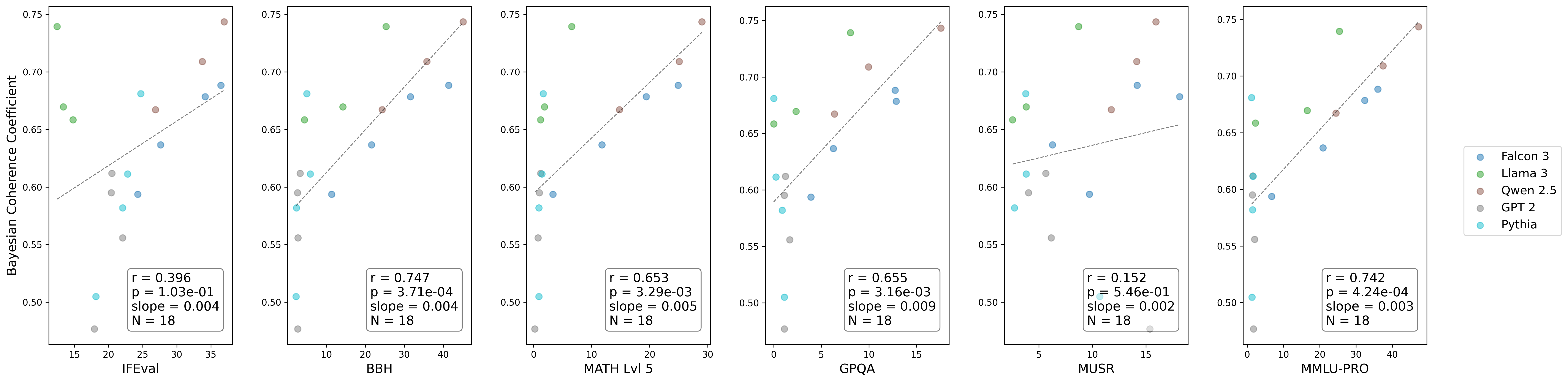}
    \caption{BCC against (normalized) scores obtained by the models on a set of benchmarks commonly used to evaluate model performance. Each point represents a pre-trained model evaluated on the full dataset. The \textit{x}-axis represents normalized benchmarks scores, with 0 as the random baseline and 100 as the maximum achievable score.}
    \label{fig:benchmarks}
\end{figure*}

\begin{table}
    \centering
   \caption{BCC, update gradient, and directional agreement between observed and expected updates for models of different sizes in the same model family. All entries are based on 6,460 evaluation instances.}
    \label{tab:model_extremes}
    \begin{threeparttable}
    \begin{tabular}{lcccc}
    \toprule
    Model&Params&BCC&Update&Direction\\
    Family&(B)&&Gradient&Agreement\% \\
    \midrule
    \multirow{2}{*}{Falcon 3} & 1.67 & 0.594 & 0.295 & 70.4 \\
     & 10.31 & 0.688 & 0.352 & 74.3 \\
    \midrule
    \multirow{2}{*}{Llama 3} & 1.24 & 0.658 & 0.381 & 73.8 \\
     & 8.03 & 0.739 & 0.457 & 74.7 \\
    \midrule
    \multirow{2}{*}{Qwen 2.5} & 3.09 & 0.667 & 0.390 & 74.3 \\
     & 14.77 & 0.743 & 0.482 & 75.8 \\
    \midrule
    \multirow{2}{*}{GPT-2} & 0.14 & 0.477 & 0.351 & 64.4 \\
     & 1.61 & 0.595 & 0.329 & 67.9 \\
    \midrule
    \multirow{2}{*}{Pythia} & 0.21 & 0.505 & 0.340 & 63.7 \\
     & 12.00 & 0.681 & 0.396 & 73.7 \\
    \bottomrule
    \end{tabular}
    \end{threeparttable}
\end{table}

\section{Discussion}

The significant positive correlation (\(r = 0.906, p < 10^{-6}\)) between the BCC and log number of model parameters (figure \ref{fig: size}) and the significant positive correlations (see figure \ref{fig:benchmarks}) between the BCC and model performance on common benchmarks provide evidence for our hypothesis that larger and more capable LLMs update their credences over propositions more consistently with Bayes' rule.

The correlations between BCC and the number of training steps (figure \ref{fig:training-data}) and between BCC and two of the six benchmarks we tested (IFEval and MUSR) are, although positive, not statistically significant for the models we tested. It is unclear whether this is due only to the limited number of models tested (and even more limited model families tested in the case of training steps) or whether there are other factors at play. It is also unclear why all models tested seem to under-update, that is, why the gradient of observed vs. expected updates is less than 1 for all models. The inverse correlation between this update gradient and the negative evidence log likelihood averaged over the class pair (see appendix, figures \ref{fig:bcc_vs_avg_evidence_log_likelihood} and \ref{fig:bcc_vs_avg_class_log_probability}) suggests that this is related to the evidences used in our analysis being a lot less likely compared to the classes.

Our results add to the body of research showing that the beliefs and preferences of larger and more capable LLMs are more logically consistent \citep{paleka_consistency_2025, mazeika_utility_2025}, while contrasting with \citet{fluri_evaluating_2023}, which found no increase in Bayesian consistency from GPT 3.5 and GPT 4, a larger and more capable model, despite finding improvements in other consistency measures. Their negative scaling results were attributed to the reversal curse \citep{berglund_reversal_2024}. If the reversal curse was indeed the reason for their negative scaling results, we would expect our scaling results to also be negative. We hypothesize that their negative results may instead have resulted from reliance on an error-based metric instead of a correlation-based metric such as BCC, which may assign lower scores to models which make more "confident" predictions. For further discussion on this, see appendix \ref{appendix: alternative_metric}.

Our results provide early evidence that larger and more capable pre-trained models update their beliefs in a manner more consistent with Bayes' rule, with BCC increasing approximately log-linearly with scale. These results have many potential implications. Larger and more capable LLMs being better modeled as Bayesian updaters indicates they may learn internally coherent models of the world, which may allow reasoning. Insofar as their internal world models are similar to those of humans, this should allow for more efficient information exchange between humans and AI systems. On the other hand, concealing information becomes more difficult as LLMs can infer the world state more accurately from subtle cues. If coupled with coherent preferences, more Bayesian agents may more closely approximate EUM behavior. Since EUMs with misaligned preferences may optimize for harmful world states, and are incorrigible by default, our results call for research in developing robust alignment and corrigibility methods.

\section{Limitations and Future Research}

Our study has some limitations. Firstly, we leave rigorous investigations into the non-significant correlations we observed (BCC with training steps, and BCC with two benchmarks) and the observed under-updating phenomena to future research.

Secondly, our analysis covers only pre-trained models and only up to 14 billion parameters. As such, caution is needed before extrapolating our results to larger and more capable models. Future research should repeat our results for larger LLMs and fine-tuned LLMs. Both instruction-tuned and reinforcement-learning-tuned models are interesting candidates for study.

Thirdly, we use cumulative token probabilities as a proxy for credence in the proposition they compose. It is unclear whether this is an accurate proxy for action-relevant belief states. Future research should investigate the validity of this assumption and explore alternative proxies for an LLM's credence in a proposition. Furthermore, evaluating BCC for untrained models with random parameter initializations can provide insight into the usefulness the metric. 

Finally, we evaluate one of many notions of coherence of beliefs, and that too with a single metric. Future research should extend our analysis to include other formulations of coherence. 

\section{Conclusion}

We hypothesized that LLMs update their credences over propositions in-context in a manner more consistent with Bayes' theorem with increasing model scale and capability. We tested this hypothesis over multiple models spanning five model families by designing a novel metric, the Bayesian Coherence Coefficient (BCC), which measures the correlation between the expected and observed updates to any evidence. Our results show significant correlations between the BCC and the log of the number of model parameters and between the BCC and model performance on benchmarks commonly used to evaluate model capability, providing evidence for our hypothesis that larger and more capable models update their credences more consistently with Bayes' theorem.

\section{Data and code availability}

The language models evaluated in this research are available for download from the HuggingFace Hub (\url{https://huggingface.co/}) and can be found by searching for their model names as they appear in the appendix, figure \ref{fig:log_odds_likelihood}. The benchmark scores and parameter counts were obtained from the Huggingface Open LLM Leaderboard 2 (\url{https://huggingface.co/datasets/open-llm-leaderboard/contents})

The dataset generated and used for evaluating the Bayesian Coherence Coefficient can be found at : \url{https://github.com/AISC10-team09/bayesian_reasoning/blob/main/data/data.json}, and the code for replicating our analysis can be found at : \url{https://github.com/AISC10-team09/bayesian_reasoning.git}

\section{Acknowledgments}

We would like to thank AI Safety Camp (\url{https://aisafety.camp/}) for bringing the authors together and supporting this project. We would also like to thank Guillaume Corlouer and the (anonymous) peer reviewers for feedback that significantly improved this research. SI would additionally like to thank Lancaster University for funding his researcher position with Professor Peter M. Atkinson.

\section{Author contributions}

\textbf{SI} proposed the original project under the supervision of \textbf{PA} and \textbf{RL}. \textbf{SI}, \textbf{IK} and \textbf{AT} implemented the code-base for the project. \textbf{AT} and \textbf{RC} generated the dataset. \textbf{SI} and \textbf{MB} analyzed the results. \textbf{SI}, \textbf{IK}, \textbf{MB}, \textbf{AT} and \textbf{RC} drafted the manuscript. \textbf{SI} addressed reviewer comments and prepared the final manuscript. \textbf{PA} and \textbf{RL} checked the methodology, and reviewed and edited the manuscript. 

\bibliography{auto_references, references}
\bibliographystyle{icml2025}

\appendix
\onecolumn

\section{Alternative Metric} \label{appendix: alternative_metric}

\subsection{Bayesian Coherence Error}

In addition to the Bayesian Coherence Coefficient (BCC) presented in the main paper, we explored an alternative error-based metric to quantify Bayesian coherence. The Bayesian Coherence Error (BCE) measures directly the deviation between the expected and observed belief updates (equation \ref{eq: BCE}). Since BCE is an error metric, smaller values should indicate greater Bayesian coherence.

\begin{equation} \label{eq: BCE}
    \begin{aligned}
    \text{BCE}(\theta, \mathcal{D}) = \frac{1}{\lvert \mathcal{D} \rvert} \sum_{(c_1,c_2,x,h,k)\in\mathcal{D}} \Big(\Delta_{\text{expected}} - \Delta_{\text{observed}}\Big)^2
    \end{aligned}
\end{equation}

where \(\Delta_{\text{expected}}\) and \(\Delta_{\text{observed}}\) are the expected and observed log odds updates defined in equations \ref{eq: expected_update} and \ref{eq: observed_update} from the main paper.

\subsection{Theoretical Limitations of BCE}

Our analysis revealed a fundamental limitation of BCE as a coherence metric. We discovered that BCE exhibits an undesirable bias toward high-entropy distributions. To illustrate this, consider the pathological case of an LLM that always outputs a uniform distribution over its vocabulary \(V\). For such a model:

\begin{equation} \label{eq: uniform_expected}
    \begin{aligned}
    \Delta_{\text{expected}} &= \text{log likelihood ratio} = \log\frac{P_\theta(x|c_1,h, k)}{P_\theta(x|c_2,h, k)} = \log\frac{\frac{1}{\lvert V \rvert}}{\frac{1}{\lvert V \rvert}} = 0
    \end{aligned}
\end{equation}

\begin{equation} \label{eq: uniform_observed}
\begin{aligned}
\Delta_{\text{observed}} &= \text{log posterior ratio} - \text{log prior ratio}
 = \log\frac{P_\theta(c_1|x,h, k)}{P_\theta(c_2|x,h, k)} - \log\frac{P_\theta(c_1|h, k)}{P_\theta(c_2|h, k)}
 = \log\frac{\frac{1}{\lvert V \rvert}}{\frac{1}{\lvert V \rvert}} - \log\frac{\frac{1}{\lvert V \rvert}}{\frac{1}{\lvert V \rvert}} = 0
\end{aligned}
\end{equation}

This results in \(\text{BCE} = 0\), suggesting perfect coherence despite the model providing no meaningful information. BCE can therefore be minimized trivially by increasing output entropy.

\subsection{Empirical Analysis}

\begin{figure}
    \centering
    \includegraphics[width=\linewidth]{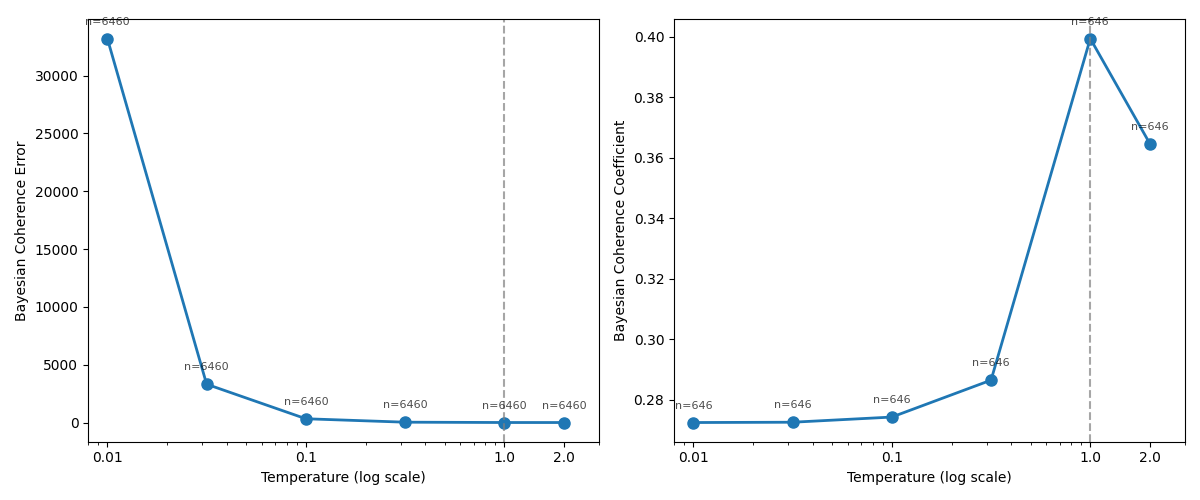}
    \caption{Temperature invariance analysis showing how BCE and BCC metrics behave under different temperature settings. Temperature scaling modifies the entropy of model outputs without changing underlying model parameters. BCC shows more robust behavior compared to BCE across different temperature values.}
    \label{fig:temperature_invariance}
\end{figure}

To empirically validate the above limitation, we analyze how the BCE and BCC metrics behave under different temperature settings (Figure \ref{fig:temperature_invariance}). Temperature scaling is a common technique that modifies the entropy of model outputs without changing the underlying model parameters.

Figure \ref{fig:temperature_invariance} confirms that correlation-based metrics like BCC are more robust measures of Bayesian coherence, as they are not confounded by the entropy of the output distribution. We use BCC as our primary metric throughout the paper.

\section{Further Results} \label{app-sec-2}

To further investigate the factors influencing Bayesian coherence and the observed under-updating phenomena (see figure \ref{fig:log_odds_likelihood}), we examine how the BCC and the update gradient vary with the average evidence log likelihood and the average class log probability, given as:

\begin{figure}
    \centering
    \includegraphics[width=\linewidth]{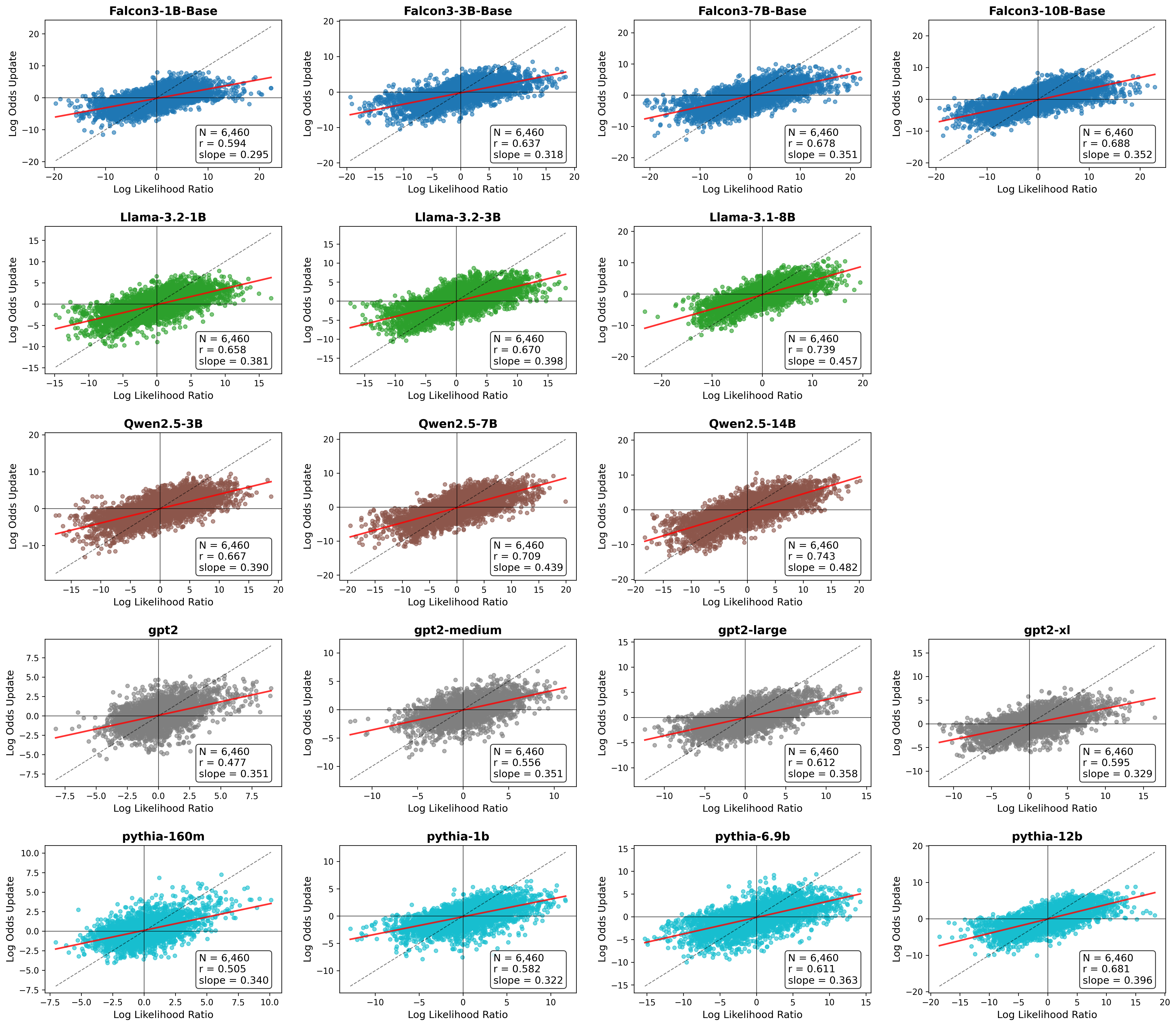}
    \caption{Scatterplots showing the observed updates (log odds
updates) against the expected updates (log likelihood ratios). Each point represents a (class pair, evidence, history, category) tuple from the dataset. The BCC of the model is the correlation (\textit{r} value) between the expected and observed updates. The solid red
and dashed black diagonal lines show the fitted and ideal update
gradients between the observed and expected updates, respectively}
    \label{fig:log_odds_likelihood}
\end{figure}

\begin{equation} \label{eq: avg_evidence_log_likelihood}
    \begin{aligned}
    \text{average evidence log likelihood} = \frac{1}{2} \big [ \log P_\theta (x|c_1, h, k) + \log P_\theta (x|c_2, h, k) \big ]
    \end{aligned}
\end{equation}

\begin{equation} \label{eq: avg_class_log_probability}
    \begin{aligned}
    \text{average class log probability} = \frac{1}{4} \bigg ( \big [\log P_\theta (c_1|h, k) + \log P_\theta (c_2|h, k) \big ] + \big [\log P_\theta (c_1|x,h, k) + \log P_\theta (c_2|x,h, k) \big ] \bigg )
    \end{aligned}
\end{equation}

\begin{figure}
    \centering
    \includegraphics[width=\textwidth]{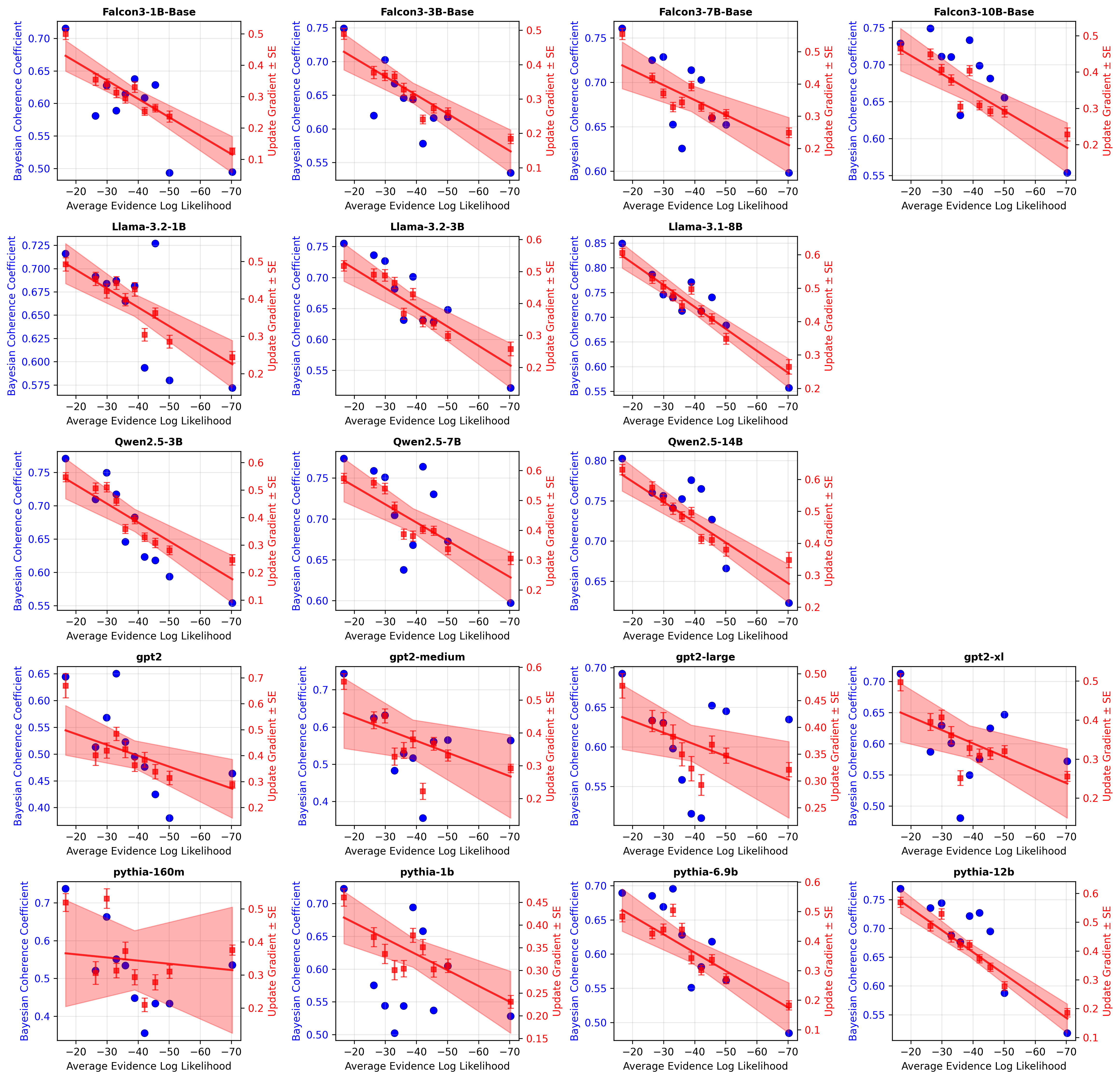}
    \caption{Scatterplots showing BCC (blue) and the gradient between the observed and expected updates (red) against the evidence log likelihood averaged over the class pair. The dataset was sorted based on the average evidence log likelihood and binned into 10 equal subsets. Each blue point represents the BCC and each red point represents the update gradient for one of 10 bins. The average evidence log likelihood is decreasing along the \textit{x}-axis.}
    \label{fig:bcc_vs_avg_evidence_log_likelihood}
\end{figure}

\begin{figure}
    \centering
    \includegraphics[width=\textwidth]{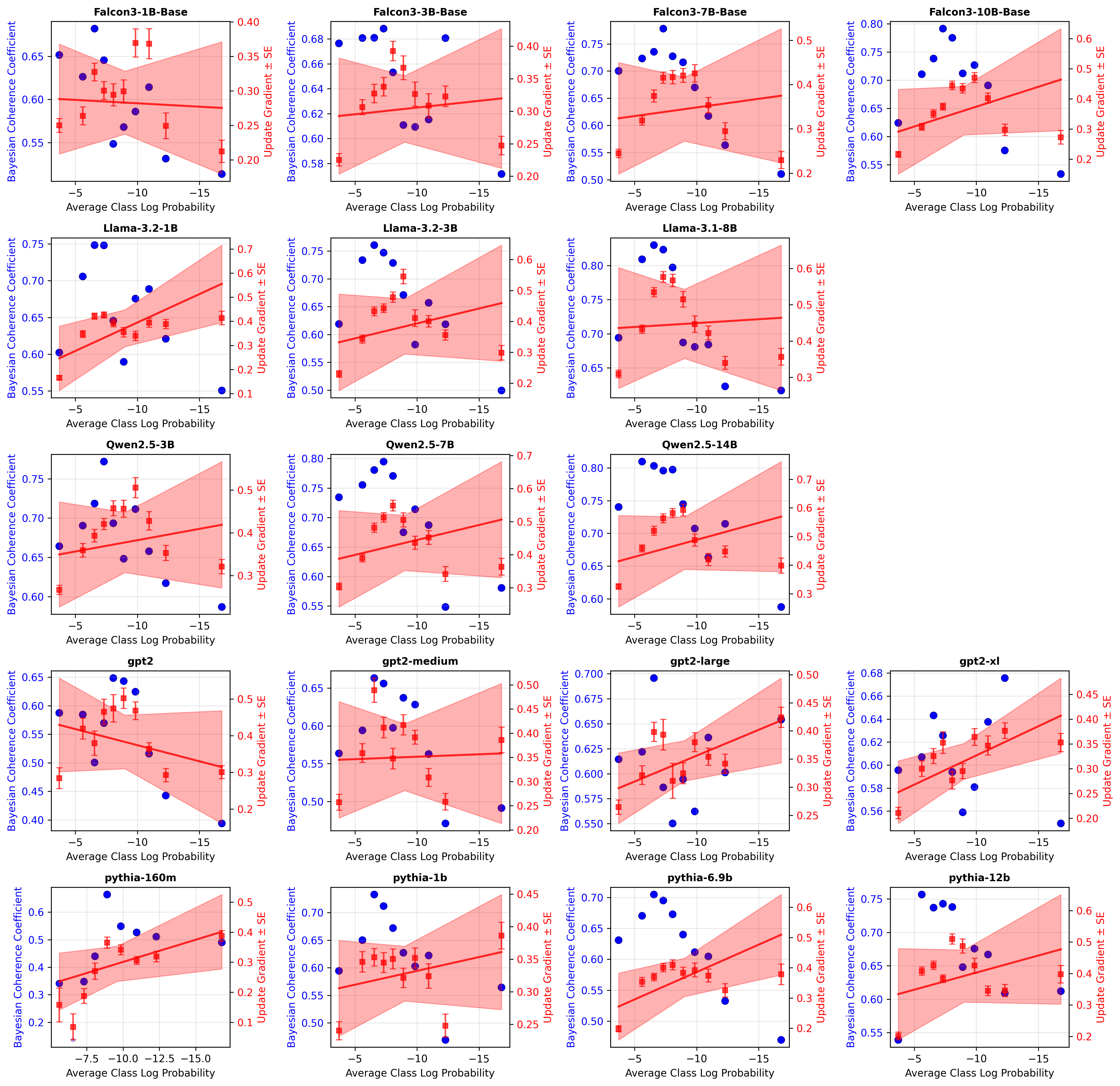}
    \caption{Scatterplots showing the BCC (blue) and the gradient between the observed and expected updates (red) against the average of the prior and posterior log probabilities averaged over the class pair. The dataset was sorted based on the average evidence log likelihood and binned into 10 equal subsets. Each blue point represents the BCC and each red point represents the update gradient for one of 10 bins. The average evidence log likelihood is decreasing along the \textit{x}-axis.}
    \label{fig:bcc_vs_avg_class_log_probability}
\end{figure}

\section{Dataset Generation} \label{app-prompt}

The prompt text (\ref{subsec:appendix_a-prompt_text}), the JSON schema (\ref{subsec:appendix_a-json_schema}) and an example (\ref{subsec:appendix_a-example}) together form the prompt given to a LLM to generate data for a specific class category.

\subsection{Desiderata}
\label{subsec:appendix_a-desiderata}
For each class category, we provide the following desiderata as instructions to the LLM:

- At least five classes in each category. \\
- All class names having the same token count. \\
- At least three conversation histories varying in how they are related to the class category, from very related to unrelated. \\
- At least 20 pieces of evidence text, distributing these pieces of evidence to favor a specific class, more than one class or none.

\subsection{Prompt Text}
\label{subsec:appendix_a-prompt_text}
\begin{tcolorbox}[colback=gray!5!white,colframe=gray!80!black,title=Prompt Text]
\begin{Verbatim}[breaklines=true,breakanywhere=true,breaksymbol=,fontsize=\footnotesize]
Data is to be generated according to the provided JSON schema. Please follow the schema exactly. There is also an example in JSON format provided. In the example the {class_category} is "novelists". Now based on the JSON schema and the example, please create data for a {class_category} "desired class category". There should be at least 5 {candidate_classes} in this category. Ensure that each of the {candidate_classes} has exactly the same number of tokens - this includes the punctuation, the space at the beginning of a class and the full stop at the end. The number of tokens should be at most 3 - use as few tokens as possible. If the {class_category} is a proper noun, then the first letter of each word of the class should be capitalized. If the {class_category} is not a proper noun then the first letter of each word should not be capitalized. There should be at least 3 {histories}, varying in how related they are to the {class_category} (from completely unrelated to very related). There should be at least 20 pieces of {evidence_text}. Some pieces of the evidence text should provide high evidence for one of the classes, other pieces of evidence text should provide evidence for several or all of the candidate classes and some pieces of evidence text should provide evidence for none of the candidate classes. Each {evidence_text} should be accompanied by an array {points_to_classes}, which is a list of classes in {class_category} that the evidence supports. This could be a single class, more than one class, all classes in the {class_category} or none (i.e. an empty list). The {evidence_elicitation} joined with the {evidence_text} should form a grammatically correct sentence including spaces and punctuation. The {class_elicitation} joined with each {class} should form a grammatically correct sentence including spaces and punctuation. It is important to follow the example for {class_elicitation} and {evidence_elicitation} including spaces and other punctuation. "desired class category" = "school_of_philosophy"
\end{Verbatim}
\end{tcolorbox}

\subsection{JSON Schema} \label{subsec:appendix_a-json_schema}
\begin{jsonschema}
{
  "schema": "https://json-schema.org/draft/2020-12/schema",
  "type": "object",
  "properties": {
    "bayesian_reasoning": {
      "type": "array",
      "items": {
        "type": "object",
        "properties": {
          "conversation_history": {
            "type": "string",
            "description": "the conversation history"
          },
          "candidate_classes": {
            "type": "array",
            "items": {
              "type": "string"
            },
			"minItems": 2,
			"uniqueItems": true,
            "description": "list of candidate classes"
          },
          "evidence": {
            "type": "string",
            "description": "justification or rationale for the classification"
          },
          "class_elicitation": {
            "type": "string",
            "description": "prompt used to elicit a candidate class"
          },
          "evidence_elicitation": {
            "type": "string",
            "description": "prompt used to elicit the evidence"
          }
        },
        "required": [
          "conversation_history",
          "candidate_classes",
          "evidence",
          "class_elicitation",
          "evidence_elicitation"
        ]
      }
    }
  },
  "required": ["bayesian_reasoning"]
}
\end{jsonschema}

\subsection{Example}
\label{subsec:appendix_a-example}



\begin{jsonschema}
{
	"bayesian_reasoning": [
		{
			"class_type": "novelists",
			"conversation_history": "We've been discussing literary styles and historical contexts in literature.",
			"candidate_classes": [" William Shakespeare.", " Oscar Wilde.", " Jane Austen.", " Charles Dickens.", " Virginia Woolf."],
			"class_elicitation": " My favourite author is",
			"evidence_elicitation": " I prefer reading",
			"evidence": [
				{
					"category": "literary_analysis",
					"evidence_text": " works that bring out the contemporary social conventions and mores of its time rather than focusing on poetic richness and dramatic performance."
				},
				{
					"category": "literary_analysis",
					"evidence_text": " character-driven narratives."
				},
				{
					"category": "historical_context",
					"evidence_text": " literature from periods of significant social transition that captures changing values, particularly those written during times when society was undergoing fundamental shifts in perspective about class, gender roles, and interpersonal relationships."
				},
				{
					"category": "historical_context",
					"evidence_text": " social observers."
				},
				{
					"category": "cultural_impact",
					"evidence_text": " books that challenged conventional thinking and introduced progressive social ideas."
				},
				{
					"category": "cultural_impact",
					"evidence_text": " enduring classics that remain relevant centuries later, particularly those that have been adapted across multiple media formats and continue to shape our understanding of narrative structure and character development in ways that transcend their original historical context."
				},
				{
					"category": "stylistic_technique",
					"evidence_text": " subtle irony."
				},
				{
					"category": "stylistic_technique",
					"evidence_text": " prose that employs wit and carefully structured dialogue to develop character."
				}
			]
		}
    ]
}
\end{jsonschema}


\end{document}